%% file: main.tex
\documentclass[letterpaper, 10 pt, conference]{ieeeconf}  

\IEEEoverridecommandlockouts                              

\usepackage{graphics} 
\usepackage{epsfig} 
\usepackage{mathptmx} 
\usepackage{times} 
\usepackage{amsmath} 
\usepackage{amssymb}  
\usepackage{booktabs}
\usepackage{gensymb}
\usepackage{multirow}
\usepackage{subcaption}
\usepackage{stfloats}
\usepackage{array}
\usepackage{siunitx} 
\usepackage{graphicx} 

\usepackage[ruled,vlined]{algorithm2e}

\newcommand{\obj}{\mathbf{o}}
\newcommand{\cam}{\mathbf{c}}
\newcommand{\body}{\mathbf{b}}

\newcommand{\pose}{\xi}

\newcommand{\Rotation}{\mathrm{R}}
\newcommand{\translation}{\mathrm{t}}
\newcommand{\scale}{\mathrm{s}}
\newcommand{\gravity}{\mathrm{g}}
\newcommand{\velocity}{\mathrm{v}}
\newcommand{\bias}{\mathrm{b}}
\newcommand{\image}{\mathcal{I}}
\newcommand{\refview}{\mathrm{ref}}
\newcommand{\disp}{\mathcal{D}}

\title{MV-ROPE: Multi-view Constraints for\\Robust Category-level Object Pose and Size Estimation}

\author{Jiaqi Yang$^{1}$*, Yucong Chen$^{1}$*, Xiangting Meng$^{1}$*, Chenxin Yan$^{1}$, Min Li$^{1}$, Ran Cheng$^{2}$, \\Lige Liu$^{2}$, Tao Sun$^{2}$, and Laurent Kneip$^{1}$
\thanks{$^{1}$ShanghaiTech University}
\thanks{$^{2}$Midea RoboZone}
\thanks{*Authors contributed equally to this work.}}

\begin{document}
\maketitle
\thispagestyle{empty}
\pagestyle{empty}

\input{sec/0_abstract}    
\input{sec/1_intro}
\input{sec/2_related_work}

\input{sec/3_methodology}
\input{sec/4_experiment}
\input{sec/5_conclusion}

{
    \small
    \bibliographystyle{IEEEtran}
    \bibliography{main}
}
\end{document}

%% file: sec/0_abstract.tex
\begin{abstract}
Recently there has been a growing interest in category-level object pose and size estimation, and prevailing methods commonly rely on single view RGB-D images.
However, one disadvantage of such methods is that they require accurate depth maps which cannot be produced by consumer-grade sensors.
Furthermore, many practical real-world situations involve a moving camera that continuously observes its surroundings, and the temporal information of the input video streams is simply overlooked by single-view methods. 
We propose a novel solution that makes use of RGB video streams.
Our framework consists of three modules: a scale-aware monocular dense SLAM solution, a lightweight object pose predictor, and an object-level pose graph optimizer.
The SLAM module utilizes a video stream and additional scale-sensitive readings to estimate camera poses and metric depth.
The object pose predictor then generates canonical object representations from RGB images. The object pose is estimated through geometric registration of these canonical object representations with estimated object depth points.
All per-view estimates finally undergo optimization within a pose graph, culminating in the output of robust and accurate canonical object poses. 
Our experimental results demonstrate that when utilizing public dataset sequences with high-quality depth information, the proposed method exhibits comparable performance to state-of-the-art RGB-D methods. We also collect and evaluate on new datasets containing depth maps of varying quality to further quantitatively benchmark the proposed method alongside previous RGB-D based methods. We demonstrate a significant advantage in scenarios where depth input is absent or the quality of depth sensing is limited.
\end{abstract}

%% file: sec/1_intro.tex
\section{Introduction}
\label{sec:intro}

The detection of objects and the estimation of their 6D pose and size is an important problem in applications such as robotics and augmented reality. Generally, the problem can be divided into instance-level and category-level pose estimation. In the former, we assume knowledge about a small number of exact shape priors (e.g. meshes, CAD models), thus reducing the problem to discrete model selection, correspondence estimation, and pose estimation.
The present work looks at the more general case of category-level pose estimation, in which the exact shape and appearance of the observed objects are assumed to be unknown. It defines the center, orientation, and size of objects at the category level, ultimately providing absolute pose and size estimations.

With the introduction of Normalized Object Coordinate Spaces (NOCS)~\cite{wang2019normalized}, there has been a surge in research efforts in recent years, continuously improving the accuracy of category-level pose and size estimation. These methods significantly broaden the applicability of object pose and size estimation in real-world scenarios.
However, while NOCS methods enable the extraction of canonical object representations from RGB images, achieving robust object pose and size estimation still requires integration with additional depth information for accurate alignment. The depth information is commonly given in the form of a direct depth channel reading or image-based depth predictions. However, it is well-understood that depth camera readings may easily suffer from measurement partiality or artifacts~\cite{camplani12holeFilling}, and that single-view depth prediction may fail to accurately reflect depth details or absolute scale~\cite{wei22depth}. Current RGB-D based methods are intrinsically limited in their real-world applicability due to their reliance on high-quality depth sensing while methods that purely rely on images are not yet able to perform on par with the state-of-the-art.

The motivation of our work is fueled by three important insights: (1) Obtaining an accurate depth map poses practical challenges for consumer-grade devices such as the iPhone or Azure Kinect, especially when it comes to small objects with smooth surfaces and intricate structures.  (2) We recognize the fact that in many practical applications, we do not only have a single image being taken of the environment, but the sensor is often mobile and continuously gathers novel views of the scene. As a result, we may indeed continuously generate predictions from nearby images and incrementally and robustly generate improved object pose predictions over time. (3) We may rely on motion stereo to perform dense depth reconstruction and thereby bypass the need for a depth camera or inaccurate single-view depth predictions.

We exploit these insights in a novel framework, denoted \textit{MV-ROPE}. It combines a scale-aware dense monocular SLAM module, a lightweight object pose estimator, as well as an object-level pose graph optimization module. Our SLAM module takes a monocular RGB image sequence along with additional scale information as input. It outputs accurate camera poses as well as dense metric depth estimations through a dense bundle adjustment layer. The lightweight object pose estimator predicts pixel-wise object canonical representations from RGB images. By performing geometric registration between depth and canonical object points, we can obtain the object pose within each keyframe. All keyframe object poses are subsequently optimized by an object-level pose graph optimization module. This allows us to ultimately obtain accurate and robust 6D object pose and size estimations. In summary, we make the following contributions:
\begin{itemize}
  \item We present the first multi-view RGB framework for robust and accurate category-level object pose estimation. Our primary modules comprise a scale-aware dense monocular SLAM module, a lightweight object pose estimator, and an object-level pose graph optimization module, which is utilized for averaging multi-view object pose estimations. The parallel operation of these modules yields robust estimations of canonical object pose and size in interactive level real-time.
  \item In an aim to circumvent the inherent scale invariance property of monocular SLAM, we propose a novel scale-aware monocular dense SLAM module making use of either inertial readings, stereo images, or direct depth readings as obtained from an RGB-D camera.
  \item Even when relying solely on images, our results demonstrate performance on par with or even better than previous state-of-the-art RGB-D methods on publicly available datasets. Additionally, we introduce a new dataset MEREAL that captures real-world scenarios, including RGB-D sequences and ground truth annotations obtained from different types and qualities of depth sensors. This dataset serves as a benchmark for evaluating the performance of our proposed method and priorly proposed RGB-D based algorithms.
\end{itemize}

We demonstrate through extensive experimental results that our algorithm achieves superior performance in situations where direct depth input is unavailable or when using low-quality depth readings.

%% file: sec/2_related_work.tex
\section{Related Works}
\label{sec:related_work}
Typical object pose and size estimation is different from object pose tracking, which is also called relative object pose estimation. Pose tracking methods~\cite{wang20206, sun2022onepose, wen2021bundletrack, wen2023bundlesdf} do not reason about an object's intrinsic shape or absolute pose, but only aim at capturing relative pose variations, primarily by tracking 3D keypoints across different frames.
On the other hand, absolute pose and size estimation of objects generally falls into one of two categories: instance-level, and category-level.
Instance-level pose estimation~\cite{xiang2018posecnn,wang2019densefusion,xu2022rnnpose}—while precise—imposes strict requirements in terms of data. It necessitates knowledge of exact CAD models or meshes corresponding to the observed objects to define each object's pose. However, securing such information is often a significant challenge in practical applications. It is not further discussed here, as the present paper focuses on category-level pose estimation. The latter expands the practical usability of object pose and size estimation in real-world scenarios. Although a few works directly learn object poses~\cite{chen2021fs,zheng2023hspose,lin2023vinet}, the fundamental idea behind category-level pose estimation is to replace CAD models with canonical object representations. By using canonical representations and real-world object points, we can directly obtain a 7DoF similarity transformation by employing geometric solvers~\cite{arun, umeyama1991least} or neural networks for implicit prediction. The process of obtaining absolute object poses can therefore be classified by the employed canonical representation:

\textbf{Deformed shape priors.} Even in the absence of an exact CAD model, a categorical shape prior can still be utilized by deforming it to match the canonical object model~\cite{tian2020shape, lin2022sar, lin2022category, deng2022icaps, zhang2022rbp, chen2021sgpa, wang2021category, meng2023kgnet, zhou2023dr,liu2023gsnet,wang2022attention}. ShapePrior~\cite{tian2020shape} learns a deformation field and applies it to a categorical shape prior, allowing for the reconstruction of the object representations in canonical space. DPDN~\cite{lin2022category} uses a self-supervised approach to generate better-deformed shape priors. Although it avoids the need for exact CAD models, it still needs representative CAD models at the category level. Recent work uses learnable queries as shape prior alternatives~\cite{wang2023query6dof} or explores the potential of prior-free methods~\cite{liu2023istnet}. 

\textbf{Generated object models.} In GCASP~\cite{li2023generative}, a latent code is utilized to reconstruct canonical object points. Additionally, Sim(3)-invariant 3D features are employed to estimate the object pose. On the other hand, works such as CenterSnap~\cite{irshad2022centersnap} and ShAPO~\cite{irshad2022shapo} utilize neural implicit representations for the canonical object model. These approaches optimize the latent code for better poses and reconstructed objects. Methods like GPV-Pose~\cite{di2022gpv} directly regress transformation to canonical space and then reconstruct object points in canonical views for refinement. These methods, however, also need geometric information from accurate depth channels for their features or latent codes to generate models.

\textbf{Regression of canonical representations.} Our method belongs to the category that utilizes semantic or geometric features to regress a canonical representation~\cite{wang2019normalized,zou20226d,chen2020learning,wan2023socs,lin2021dualposenet}. NOCS~\cite{wang2019normalized} utilizes a single RGB image to predict normalized object coordinates for each pixel. It then employs correspondences between NOCS points and object points from the depth image to get object pose and size.
MetricScale~\cite{lee2021category} independently predicts object metric scale and object center by networks and utilizes predicted NOCS to recover object orientations.
~\cite{wan2023socs} enhances NOCS by semantically-aware keypoints guidance, which requires additional depth input. Regression of canonical representations is intuitive and easy to operate. More importantly, such methods have the potential to use the semantic information contained in RGB images to obtain geometric information, which avoids the need for direct depth input channels. 

To the best of our knowledge, our method is the first to employ multi-view constraints of input RGB video streams to obtain reliable absolute object pose estimation results.

%% file: sec/3_methodology.tex
%

\section{Methodology}
\label{sec:methodology}
We will first provide an overview of the system before going into further details on camera pose and dense depth estimation, object instance segmentation and association, object pose estimation, and pose graph optimization.

\begin{figure*}[t]
  \centering
  \includegraphics[width=\textwidth]{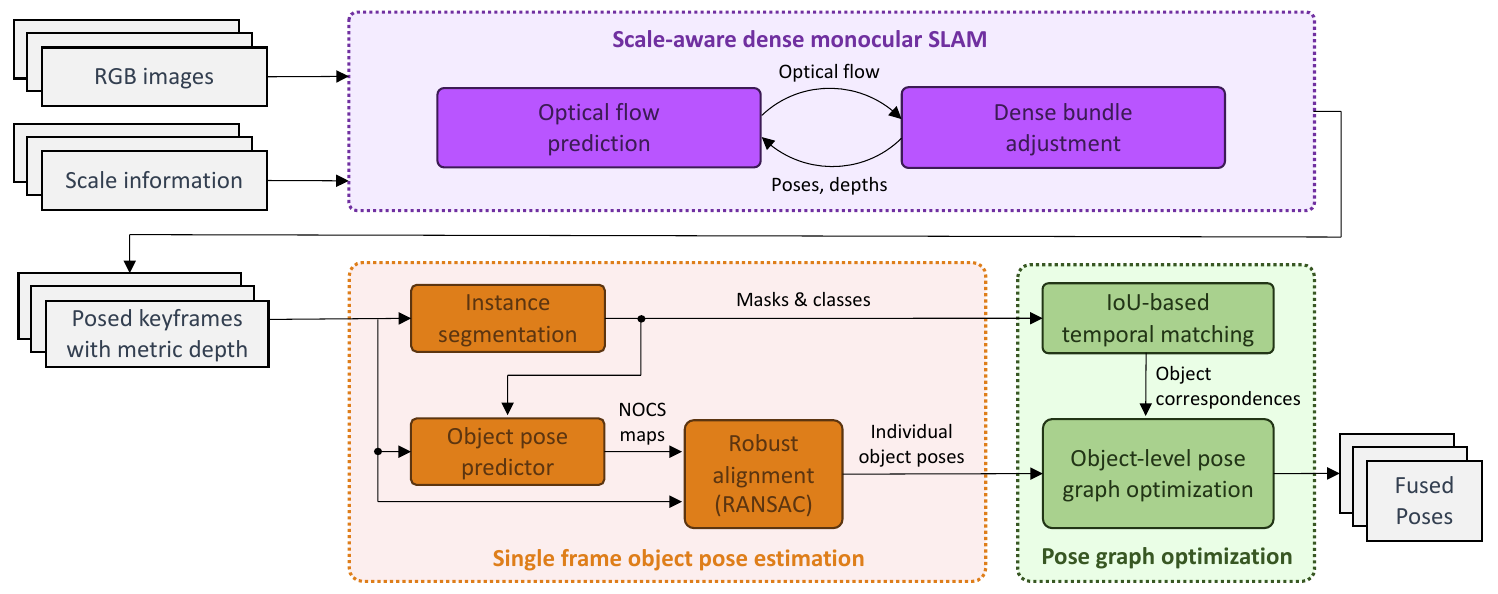}
  \caption{Overview of the complete, proposed system. The input for the first block (in purple), is a continuous image stream accompanied by scale information. The output of this block includes keyframe camera poses and dense metric depth maps. 
  The second block (in orange) is the single-view object pose estimation module. It utilizes an object pose estimator to obtain NOCS maps for each segmented object. These NOCS maps are then aligned with partial dense depth maps using a RANSAC framework to estimate the pose of each individual object. A third block (in green) finally establishes object correspondences and optimizes object poses over time.}
  \label{fig:overview}
\end{figure*}
\subsection{Overview}
\label{sec:overview}
From a high-level perspective, our framework accepts an RGB image sequence $\{\image_i\}$ and one of a few possible scale-sensitive readings. Additional information to help resolve the global scale factor can be in the form of stereo images or merely IMU measurements. For comparison purposes, a third alternative using direct depth readings is also supported. The output of our framework consists of camera poses and dense metric depth maps for each keyframe $i$, represented as $\pose_{\refview}^{c_i}\in\mathrm{SE}(3)$ and $\disp_i\in\mathbb{R}^{H\times W}$, respectively. Additionally, our framework provides the pose of each object instance $k$ within a global reference frame, denoted as $\pose_{\refview}^{\obj_k}\in \mathrm{Sim}(3)$. To facilitate comprehension, we designate the first keyframe as the reference frame, and $\image_\refview = \image_{0}$.

For ease of representation, going forward, we will consistently use superscript $\obj$ to represent objects, $\body$ to represent the IMU body,  and superscript $\cam$ to represent the camera frame. We use $\Rotation$ and $\translation$ to represent the rotation and translation part of pose $\pose$. Additionally, we will use indices $i$ and $k$ to refer to frames and objects, respectively.

Our framework is illustrated in Figure~\ref{fig:overview}. The first block runs scale-aware dense monocular SLAM with the purpose of obtaining accurate poses and metric depth maps for each keyframe (details in Sec.~\ref{sec:pose_and_depth}). Towards object-centric perception, our system includes a second block that consists of an instance segmentation network and a lightweight object predictor. Inspired by ~\cite{wang2019normalized}, the object pose predictor predicts NOCS maps for each individual object instance. Having NOCS maps and the back-projected 3D points derived from the depth maps, the second, single-view block concludes by employing the Umeyama algorithm \cite{umeyama1991least} within RANSAC to compute individual object poses. Concurrently, a third block takes all single view object pose estimates and incorporates them along with keyframe poses into a multi-view object-level pose graph optimization module. The entire process results in incrementally refined, globally consistent canonical object poses $\pose^{\obj_k}_{\refview}$ for each instance $\obj_k$. 

\begin{figure}[b]
  \centering
  \includegraphics[width=\linewidth]{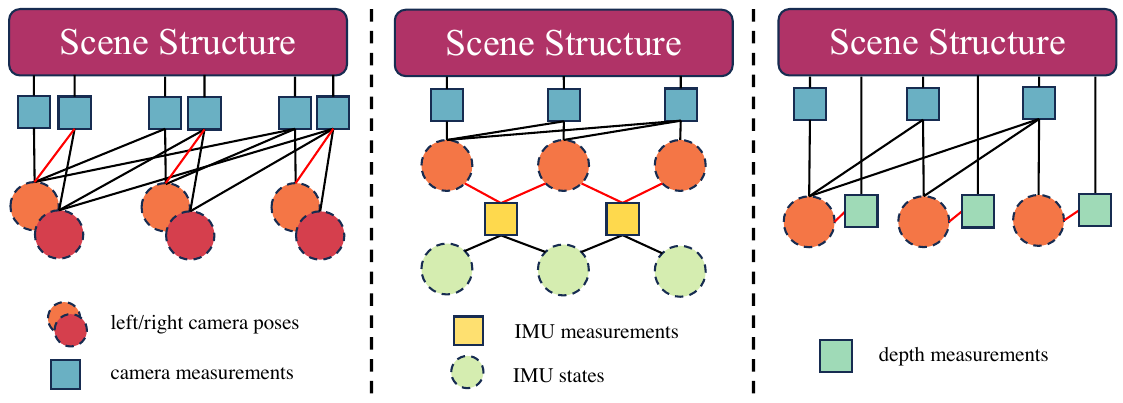}
  \caption{Graphical models of our scale-aware dense SLAM system. From left to right: model using stereo images, IMU measurements, and depth readings. The red lines represent scale-aware factors in our bundle adjustment layer.}
  \label{fig:scale_factor}
\end{figure}
\subsection{Camera Pose and Metric Depth Estimation}
\label{sec:pose_and_depth}
Our camera pose and metric depth estimation module is inspired by other dense geometric bundle adjustment frameworks such as DROID-SLAM~\cite{teed2021droid}, and VOLDOR++~\cite{min21voldor}, and consists of two elements: recurrently refined optical flow and a scale-aware dense bundle adjustment layer. The optical flow module is designed based on~\cite{teed2020raft}, and the accompanying scale-aware dense bundle adjustment layer optimizes both pose and depth while incorporating scale constraints from different sources of scale information.
The objective includes a reprojection error term and a scale constraint term.
\subsubsection{Reprojection Error Minimization}
The first objective aims to minimize the covariance reweighted sum of squared reprojection errors. The reprojection errors are calculated as the difference between the target location of 2D image points as predicted by optical flow and as obtained by applying the estimated relative pose and the dense depth to perform image warping. The reprojection objective is given by:
\begin{equation}
    E_{r}(\pose_{\refview}^{\cam}, \disp) = \sum_{\{i,j\}\in \mathcal{E}}||\mathcal{P}_{ij}- \Pi_{\cam}\left(
    \tau\left(\pose^{\cam_j}_{\cam_i}, \Pi_{\cam}^{-1}(\disp_i)\right)\right)||^{2}_{\sum_{ij}},
\end{equation}
where $\pose_{\refview}^{\cam}=\{\pose_{\refview}^{\cam_i}\}$ is the set of all keyframe poses to be estimated, $\disp=\{\disp_i\}$ is the set of all dense depth maps to be estimated, $\mathcal{E}$ is the set of all keyframe-pairs between which residuals are evaluated, $\Pi_{\cam}$ is a camera projection function that takes a $H\times W\times 3$ tensor of 3D world points and returns the $H\times W\times 2$ tensor of corresponding image points, and $\Pi_{\cam}^{-1}$ is the corresponding inverse mapping that takes a dense depth field $\disp_i$ and returns the corresponding $H\times W\times 3$ tensor of 3D points expressed in the camera frame. $\pose^{\cam_j}_{\cam_i}=\pose^{\cam_j}_{\refview}(\pose^{\cam_i}_{\refview})^{-1}$ is the euclidean transformation from frame $i$ to $j$, and $\tau(\cdot,\cdot)$ is a function defined to take a euclidean transformation and a $H\times W\times 3$ tensor of 3D world points, and return the equal-size tensor of transformed 3D world points. Finally, $\mathcal{P}_{ij}$ is the target location of the points in frame $i$ in frame $j$ as hypothesized by the optical flow prediction between frames $i$ and $j$~\cite{teed2020raft}, and $\sum_{ij}$ expresses the uncertainty of these predictions. 

\subsubsection{Scale Constraints}
For scale-aware bundle adjustment, please refer to Figure.\ref{fig:scale_factor}.
When the scale input is in the form of stereo images, we can input both the left and right camera images into the bundle adjustment layer. At the same time, the extrinsics of the left and right cameras 
$\pose^{\cam'}_{\cam}$ are known and fixed. The right camera is marked with a prime, and the objective is given by
\begin{equation}
    E=E_r+\sum_{\{i,i'\}\in \mathcal{E}}||\mathcal{P}_{ii'}- \Pi_{\cam}\left(
    \tau\left(\pose^{\cam'}_{\cam}, \Pi_{\cam}^{-1}(\disp_i)\right)\right)||^{2}_{\sum_{ii'}}.
\end{equation}

When the scale input is in the form of IMU measurements, we use visual-inertial alignment as introduced in ~\cite{qin2018vins} to recover metric scale of visual SLAM estimates and initialize IMU gravity $\gravity$, velocity $\velocity$ and bias $\bias$. 
After a successful initialization, the vision reference frame can be aligned with the inertial frame, and rescaled based on the identified velocity.
The continuous constraint of scale and inertial alignment is then realized by the addition of IMU pre-integration terms. The latter are finally used to form inertial residuals between consecutive keyframes as given by
\begin{align}
    e_{\body_i}&=||\Rotation^{\body_{i}}_{\refview}(\translation^{\refview}_{\body_{i+1}} - \translation^{\refview}_{\body_{i}} +\frac{1}{2}\gravity\Delta t_{i}^2-\velocity_{\body_{i}}^{\refview}\Delta t_{i})-\hat\alpha^{\body_{i+1}}_{\body_{i}}||^2 \\
    &+||\Rotation^{\body_{i}}_{\refview}(\velocity_{\body_{k+1}}^{\refview} - \gravity\Delta t_{i}) - \hat\beta^{\body_{i+1}}_{\body_{i}}||^2 \\
    &+||\mathrm{log}({\Rotation_{\body_{i}}^{\refview}}^{-1}\Rotation_{\body_{i+1}}^{\refview}\hat\gamma^{\body_{i+1}}_{\body_{i}})||^2\\
    &+||\bias^{a}_{i}-\bias^{a}_{i+1}||^2+||\bias^{g}_{i}-\bias^{g}_{i+1}||^2,
\end{align}
where $\hat\alpha,\hat\beta,\hat\gamma$ are the measured translation, velocity and rotation changes between keyframe IMU states $\body_{i}$ and $\body_{i+1}$ as given by IMU pre-integration. $\Delta t_i$ is the time interval.
Visual-inertial bundle adjustment differs in that it also optimizes the keyframe IMU states ${\body_i}$ consisting of translational velocity and gyroscope and accelerate-meter biases. The objective now reads $E=E_r+\sum_{i\in\mathcal{V}}e_{\body_i}$, where $\mathcal{V}$ is the set of keyframes in the bundle adjustment layer.

In the case where the scale input is in the form of a depth map, it is possible to recover the correct scale by directly incorporating a constraint term into the estimated depth map.
The final scale-aware objective is given by
\begin{equation}
    E=E_r+\sum_{i\in\mathcal{V}} ||\hat{\mathcal{D}_i}-\mathcal{D}_i||^2,
\end{equation}
where $\hat{\mathcal{D}_i}$ is the depth measurements from the depth camera. Note that the inclusion of direct depth readings is not our main focus, but the objective is used for comparative purposes.

We use Gauss-Newton-style optimization methods to solve these nonlinear optimization objective, and---in analogy to bundle adjustment methods---use the Schur complement trick to accelerate the computation.
\subsection{Single View Object Pose Estimation}
\label{sec:single_view_object_pose}
\subsubsection{Instance Segmentation and Object Association}
\label{sec:object_tracking}
We can use off-the-shelf toolbox like SAM-based models~\cite{cheng2023segment,liu2023grounding} or Mask-RCNN~\cite{he2017mask} for instance segmentation. 
To build associations between each object, we then calculate the Intersection over Union (IoU) across all instances of each two frames. If two bounding boxes have the same category and the IoU exceeds a certain threshold, the two object instances are assigned to the same instance ID.

\subsubsection{Object Pose Predictor}
\label{sec:nocs_network}
Our object pose predictor is built upon NOCS~\cite{wang2019normalized} and aims to produce canonical representations of objects. It processes RGB keyframes and incorporates object masks and class labels from our instance segmentation and tracking module. The output comprises NOCS maps for each object, serving as a crucial input for the subsequent object pose estimation module.
The fundamental design of our NOCS predictor is inspired by the original NOCS implementation~\cite{wang2019normalized}. However, we have already acquired the necessary masks and class labels, thereby eliminating the need for the Mask R-CNN-like framework~\cite{he2017mask}. Instead, we leverage an encoder adapted from the U-Net architecture~\cite{ronneberger2015u} to attain a feature map for each complete, full-resolution input keyframe. This step is followed by using the object masks to crop the ROI features and resize them to a standard $32\times32$ grid. Subsequently, three separate predictors are engaged to acquire the $x$, $y$, and $z$ components of the NOCS output.

\subsubsection{Robust Geometric Registration for Object Pose}
For each masked object, its pose is estimated by registering the NOCS points $X^{nocs}\in \mathbb{R}^{3\times N}$ and the back-projected object depth points $X^{xyz} \in \mathbb{R}^{3\times N}$ using Umeyama's algorithm~\cite{umeyama1991least}. Note that the correspondences between $X^{nocs}$ and $X^{xyz}$ are simply given by the question whether or not they originate from the same pixel in the image. In order to deal with inaccurate instance segmentations and noisy NOCS predictions, we apply RANSAC for outlier rejection. The fitted model is the similarity transformation
\begin{equation}
    X^{xyz} = s \Rotation X^{nocs}+\translation,
\end{equation}
where $\scale\in\mathbb{R}$ is the scale of the object, $\Rotation \in \mathrm{SO}(3)$ is the orientation of the object, and $t\in\mathbb{R}^3$ its position. RANSAC with post-refinement over all inliers hence aims at a robust minimization of the energy objective
\begin{equation}
    \scale^*,\Rotation^*,\translation^* = \underset{\scale, \Rotation, \translation}{\arg\min}\sum_{i}^N||X^{xyz}_i -\scale \Rotation X^{nocs}_i-\translation||^2.
\end{equation}
\subsection{Object-Level Pose Graph Optimization}
\label{sec:pose_graph_optimization}
One of our assumptions is that the object is static. Therefore, the camera pose and object pose can be optimized through the proposed object-level pose graph. Our graph contains where $K$ nodes representing object poses and $N$ node representing camera poses. There are two types of edges in our pose graph. The first one is camera-camera edges, which we can obtain from our scale-aware dense monocular SLAM-based relative pose estimations $\pose_{\cam_{i}}^{\cam_{i+1}}$. Its corresponding covariance matrix $\Sigma_{\cam_{i}}$ is derived from the bundle adjustment layer. The second type is given by object-camera edges. The relative pose between object node $k$ and camera node $i$ is denoted $\pose_{\obj_k}^{\cam_i}$, which is the individual object pose obtained from each keyframe.
The covariance matrix of the object camera edges can be approximated as
\begin{equation}
    \Sigma_{\obj_{ki}} = (J_{\obj_{ki}}^T\cdot J_{\obj_{ki}})^{-1},
\end{equation}
where $J_{\obj_{ki}}$ is the jacobian matrix of the estimated object pose, defined as 
\begin{equation}
    J_{\obj_{ki}}=\frac{\partial{({\pose^{\obj_k}_{\cam_i} \cdot X_{\obj_{ki}}^{nocs}}-X_{\obj_{ki}}^{xyz})}}{\partial \pose^{\obj_k}_{\cam_i}}|_{\pose^{\obj_k}_{\cam_i} = {\pose^{\obj_k}_{\cam_i}}^*},
\end{equation}
where $X_{\obj_{ki}}^{xyz}$ and $X_{\obj_{ki}}^{nocs}$ are back-projected depth points and predicted NOCS points of object $\obj_k$ in frame $i$.
Then our object-level pose graph optimization problem can be formulated as:
\begin{equation}
    \underset{\{\pose^{\obj_{k}}_{\refview}\},\{\pose_{\refview}^{\cam_i}\}}{\arg\min}\sum_{k}^{K}\sum^{N}_{i} e_{k,i}^T\cdot\Sigma^{-1}_{\obj_{ki}}\cdot e_{k,i}+\sum^{N-1}_{i}e^T_{i,i+1}\cdot\Sigma^{-1}_{\cam_{i}}\cdot e_{i,i+1},
\end{equation}
where $e_{k,i}=\mathrm{log}(\pose^{\obj_k}_{\refview}(\pose^{\cam_i}_{\refview})^{-1}\pose^{\cam_i}_{\obj_{k}})$ and $e_{i,i+1}=\mathrm{log}(\pose^{\cam_i}_{\refview}(\pose^{\cam_{i+1}}_{\refview})^{-1}\pose_{\cam_{i+1}}^{\cam_i})$ are the residual terms derived from camera-object edge and camera-camera edges, respectively.

We can use robust loss functions like a Huber loss to reduce the influence of outlier object poses. Our implementation is based on ~\cite{rosen2019se}, which can produce certifiably globally optimal results to our object-level pose graph optimization problem in the presence of outlier object-camera edges.

%% file: sec/4_experiment.tex
\section{Experiments}
\label{sec:experiments}
We now proceed to our experimental evaluation. We start by introducing implementation details, and discuss our baseline methods and the benchmarks used for evaluating the 6DoF pose and size estimations. Later, we will introduce details of a self-collected dataset named \textit{MEREAL} (MV-ROPE Extended Real dataset), which is used to benchmark state-of-the-art methods as well as our proposed method with varying qualities of depth sensing. While the detailed quantitative performance of our proposed method is show-cased on public datasets, we also provide qualitative and quantitative results on MEREAL, ultimately revealing the superiority of the proposed method.

\subsection{Implementation Details}
\label{sec:implementation_details}
In our scale-aware dense monocular SLAM module, the recurrent optical flow estimator and dense bundle adjustment layer are inspired by the design of DROID-SLAM~\cite{teed2021droid}. Building upon this foundation, we extended the additional inputs to include depth images, stereo images, or IMU measurements, encompassing various forms of scale information. Furthermore, we seamlessly integrated these pieces of information into the bundle adjustment layer. Note that we employ its pre-trained weights of ~\cite{teed2021droid} for the optical flow estimation without performing any fine-tuning. 
The object pose predictor is trained on the NOCS dataset \cite{wang2019normalized}. The dataset consists of two splits: CAMERA and REAL. The CAMERA split is generated by rendering synthetic objects into 300k real-world images. The Real split contains RGB-D image sequences from 31 indoor scenes. Our training strategy aligns with the policies set forth by ShAPO \cite{irshad2022shapo}. Initially, the model undergoes training on the CAMERA split, followed by fine-tuning on the REAL training split. To address the challenge of rotational symmetries, the loss function is designed to rotate symmetric objects (bottle, bowl, and can) around the y-axis of the predicted NOCS coordinates. It then selects the smallest loss from rotations along the circle. This ensures a more accurate and reliable model. The encoder is structured as a modified U-Net, where the last two up-sampling blocks are excluded. Instead, we utilize $4\times$interpolation to regain the full resolution. This modification is crucial to maintain the channel depth, thereby ensuring that the feature encapsulates sufficient information. Following the encoder, each of the three separate predictors is a shallow CNN composed of five convolutional layers. This design choice contributes to the efficiency and compactness of the model. Overall, the object pose predictor is an efficient, lightweight solution and can be trained on a single NVIDIA GeForce RTX 2080 Ti. Additionally, the core module of our framework achieves real-time performance at an interactive level, with a frame rate of about 5 keyframes per second.
\begin{figure}[tbp]
\centering
    \begin{subfigure}[b]{0.155\textwidth}
        \includegraphics[width=\linewidth]{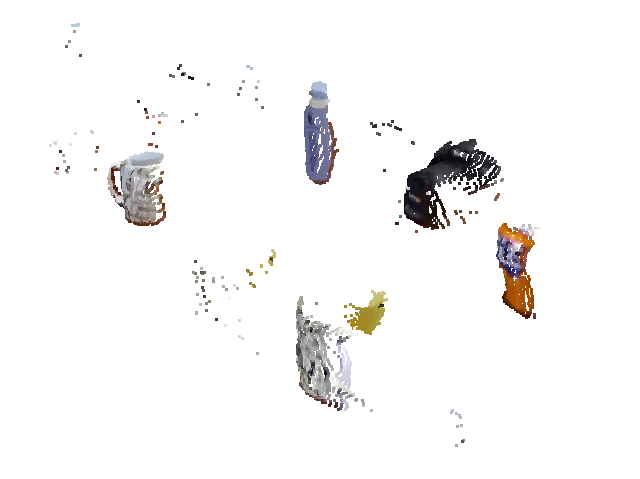}
        \caption{ZED}
        
    \end{subfigure}
    \begin{subfigure}[b]{0.155\textwidth}
        \includegraphics[width=\linewidth]{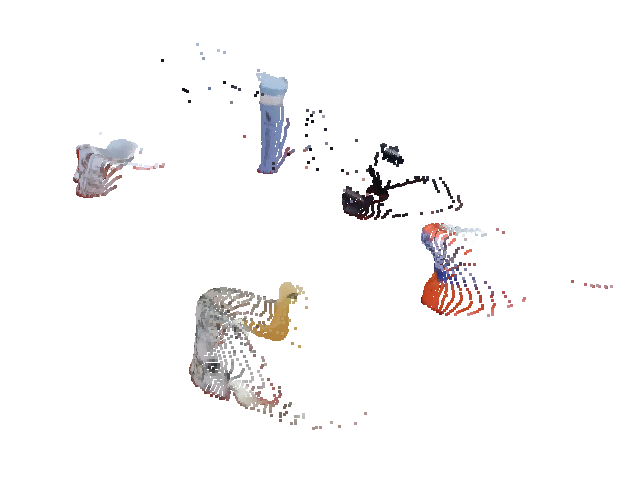}
        \caption{Azure Kinect}
        
    \end{subfigure}
    \begin{subfigure}[b]{0.155\textwidth}
        \includegraphics[width=\linewidth]{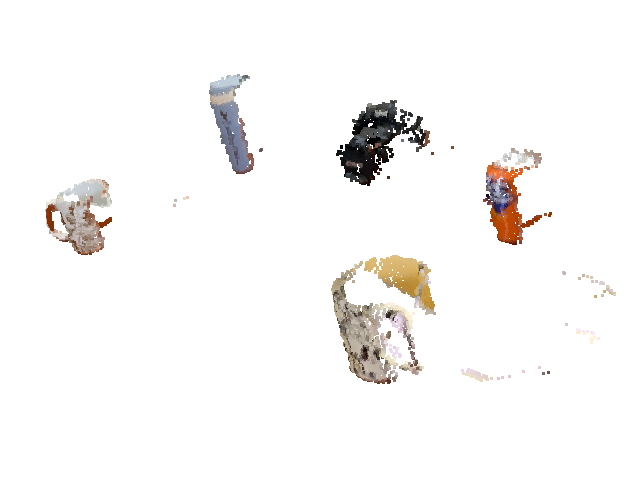}
        \caption{Structure Sensor}
        
    \end{subfigure}
    \caption{Illustration of object point clouds captured by different sensors.}
    \label{fig:sceneMREAL}
\end{figure}  
\subsection{MEREAL Dataset}
\label{sec:MEREAL_dataset}
This dataset comprises sequences recorded using different depth sensors, and offers depth images and IMU measurements.
Our depth sensors include ZED, Azure Kinect, and Structure Sensor. ZED is a stereo camera that has undergone precise calibration for both intrinsic and extrinsic parameters. It can also output depth maps through stereo reconstruction algorithms. Azure Kinect is a commonly used consumer-grade RGB-D camera that utilizes Time-of-Flight (ToF) technology for depth sensing. It also incorporates a hardware-synchronized IMU sensor. Structure Sensor is a high-precision 3D scanner and the same device used in the NOCS REAL dataset. It is capable of generating high-quality depth maps. Furthermore, Structure Sensor can be used in conjunction with an iPad, allowing for the joint calibration of the depth camera and iPad's color camera to obtain well-aligned RGB-D data. As presented in Figure.~\ref{fig:sceneMREAL}, we can observe that the Azure Kinect has inaccurate depth, while ZED has substantial noise.

Our dataset comprises three different scenes, each with different objects and object placements. During the recording process, each scene was captured 3 times using each of the different cameras. As a result, we have a total of 27 sequences, with each sequence containing 500-1000 frames.
In each scene, we have an additional calibration board placed to acquire ground truth camera poses. 
Before recording the data, we perform 3D reconstruction for each object in the scene to obtain accurate object models. 
Regarding the pose annotation of the object, we will manually align the back-projected point cloud of the object to the reconstructed object model for the object pose.
Furthermore, we only need to first annotate the object pose in the first frame. Then, we can propagate the object pose from the first frame to all the subsequent frames using the camera motion obtained from the calibration board.

\begin{table}[htbp]
    \centering
    \resizebox{0.48\textwidth}{!}{%
        \begin{tabular}{ccccccccc}
            \cmidrule[1pt]{1-9}
            \textbf{Method} & 
            \textbf{Extra inputs} & \textbf{IoU25} &
            \textbf{IoU50} & \textbf{IoU75} & \textbf{5\degree{}}2cm & \textbf{5\degree{}}5cm & \textbf{10\degree{}}2cm & \textbf{10\degree{}}5cm \\
            \cmidrule(lr){1-1}\cmidrule(lr){2-9}
            NOCS\cite{wang2019normalized} & \textbf{d} & 84.8 & 78.0 & 30.1 & 7.2 & 10.0&  13.8 & 25.2  \\

            DPDN\cite{lin2022category} & \textbf{d+p} & -& 83.4 & 76.0 & 46.0 & 50.7 & 70.4 & 78.4   \\
            GPV-Pose\cite{di2022gpv} & \textbf{d} & 84.2& 83.0  & 64.4 & 32.0 & 42.9& - &73.3   \\
            VI-Net\cite{lin2023vinet} & \textbf{d} & - & - & 48.3 & \textbf{50.0} & \textbf{57.6} & 70.8 & 82.1 \\
            Query6DoF\cite{wang2023query6dof} & \textbf{d} & - & 82.9 & 76.0 & 46.8 & 54.7 & 67.9 & 81.6 \\
            HS-Pose\cite{zheng2023hspose} & \textbf{d} & 84.2 & 82.1 & 74.7 & 46.5 & 55.2 & 68.6 & \textbf{82.7} \\
            IST-Net\cite{liu2023istnet} & \textbf{d} & 84.3 & 82.5 & \textbf{76.6} & 47.5 & 53.4 & \textbf{72.1} & 80.5 \\
            SOCS\cite{wan2023socs} & \textbf{d} & - & 82 & 75 & 49 & 56 & 72 & 82 \\
         \cmidrule[0.5pt](lr){1-9}
         \textbf{Ours}& \textbf{s} & \textbf{99.9} &\textbf{93.6} & 60.3 & 31.4 & 46.6 & 46.1 & 73.7\\
         \cmidrule[1pt](lr){1-9}
        \end{tabular}%
    }
    \caption{Quantitative comparisons of mAP on REAL\cite{wang2019normalized}. We marked the extra inputs of each method to distinguish their technical routes. \textbf{d} means depth images, \textbf{p} means categorical shape priors, and \textbf{s} means scale information. For the metrics, IoUx means mAP defined by 3D IoU over a threshold of x\%; \textbf{m\degree{}}n\,cm represents mAP defined by rotation error less than n\degree{} and transformation error less than m cm.}
    \label{table:quantitative_real275}
\end{table}

\subsection{Results on NOCS Dataset}
\label{sec:quantitative_REAL}
Our method uses RGB-D image sequences as an input on this dataset. The CAMERA split of the NOCS dataset~\cite{wang2019normalized} consists of single view RGB-D images, only, hence we conducted our experiments on the REAL test split. Regarding baseline methods, we selected state-of-the-art approaches ~\cite{di2022gpv,liu2023istnet,lin2023vinet,zheng2023hspose,wang2023query6dof,wan2023socs}, from the three different categories as discussed in the related work section (Sec.~\ref{sec:related_work}). 

\begin{figure}[htbp]
\centering
    \begin{subfigure}[b]{0.15\textwidth}
        \includegraphics[width=\linewidth]{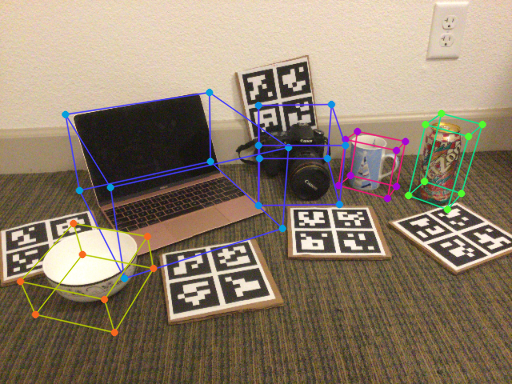}
    \end{subfigure}
    \begin{subfigure}[b]{0.15\textwidth}
        \includegraphics[width=\linewidth]{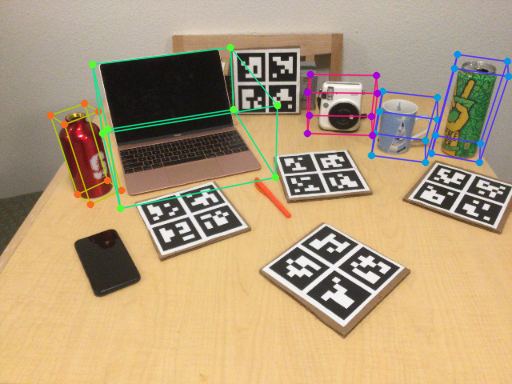}

    \end{subfigure}
    \begin{subfigure}[b]{0.15\textwidth}
        \includegraphics[width=\linewidth]{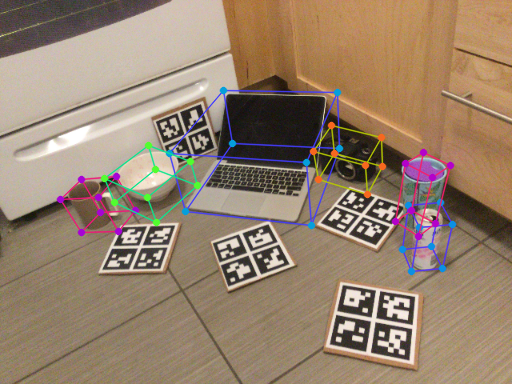}

    \end{subfigure}
    \begin{subfigure}[b]{0.15\textwidth}
        \includegraphics[width=\linewidth]{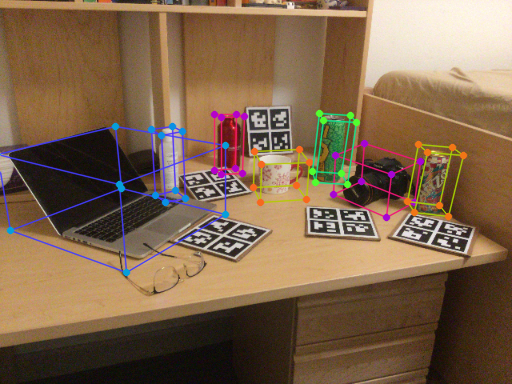}

    \end{subfigure}
    \begin{subfigure}[b]{0.15\textwidth}
        \includegraphics[width=\linewidth]{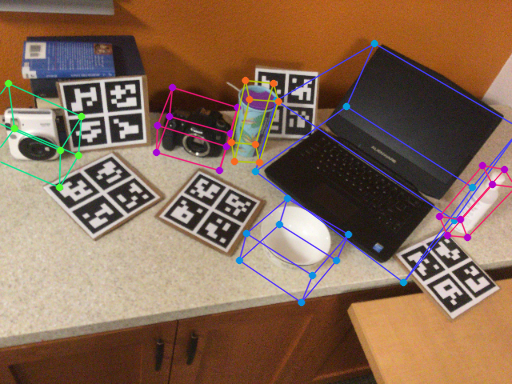}

    \end{subfigure}
    \begin{subfigure}[b]{0.15\textwidth}
        \includegraphics[width=\linewidth]{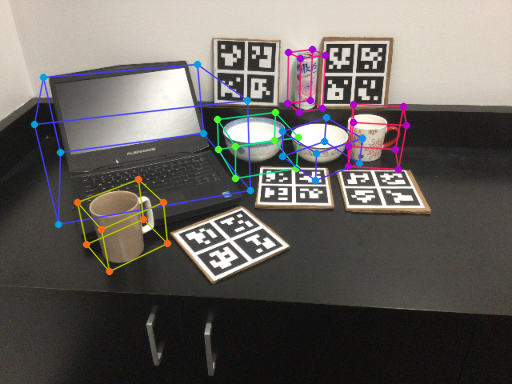}

    \end{subfigure}
    \caption{Qualitative results of our method obtained on all 6 sequences of the REAL test dataset~\cite{wang2019normalized}.}
    \label{fig:qualitativeResultsREAL}
\end{figure}

\begin{figure}[htbp]
    \begin{subfigure}[b]{0.15\textwidth}
        \includegraphics[scale=0.18]{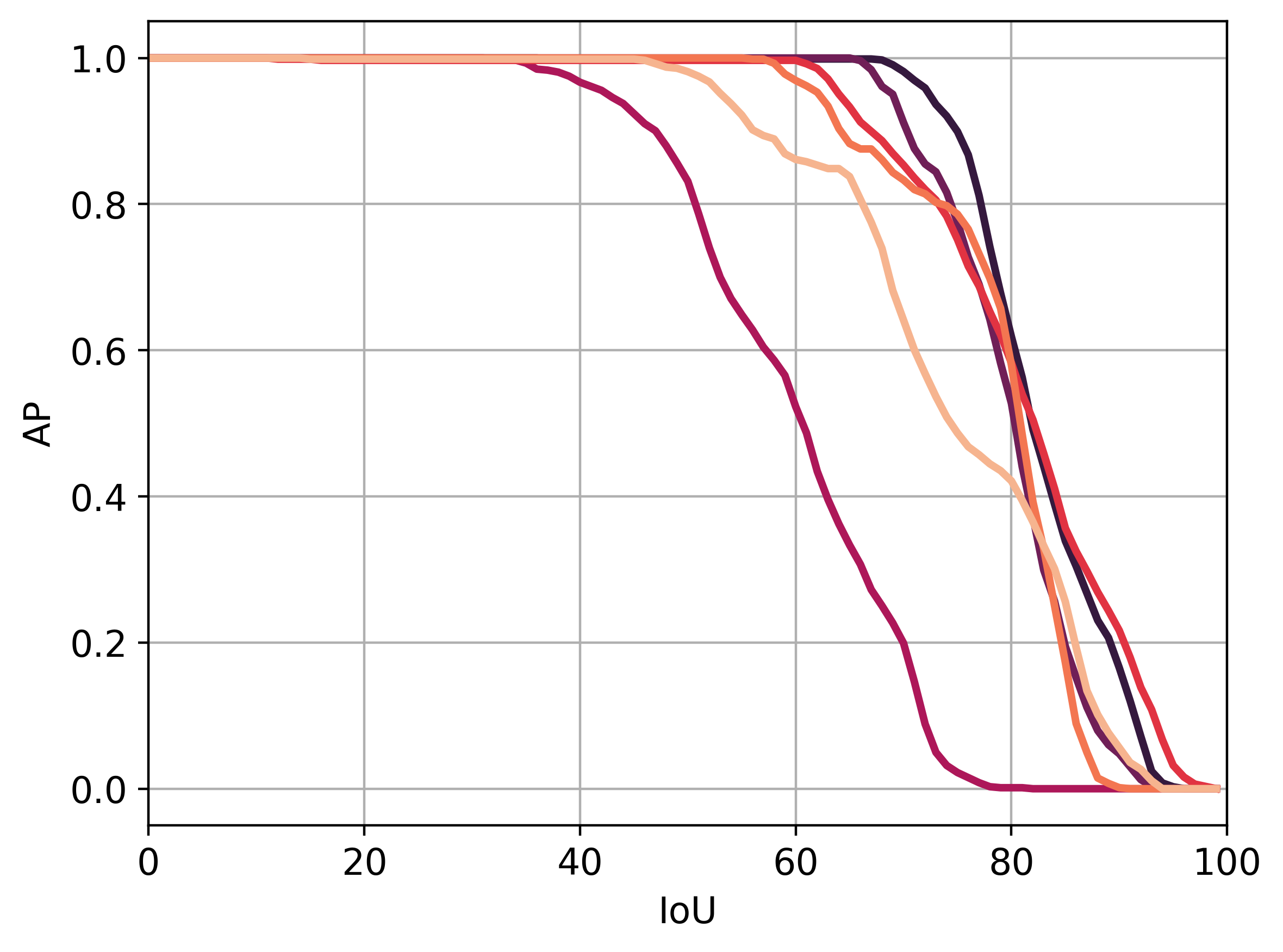}
    \end{subfigure}
    \begin{subfigure}[b]{0.15\textwidth}
        \includegraphics[scale=0.18]{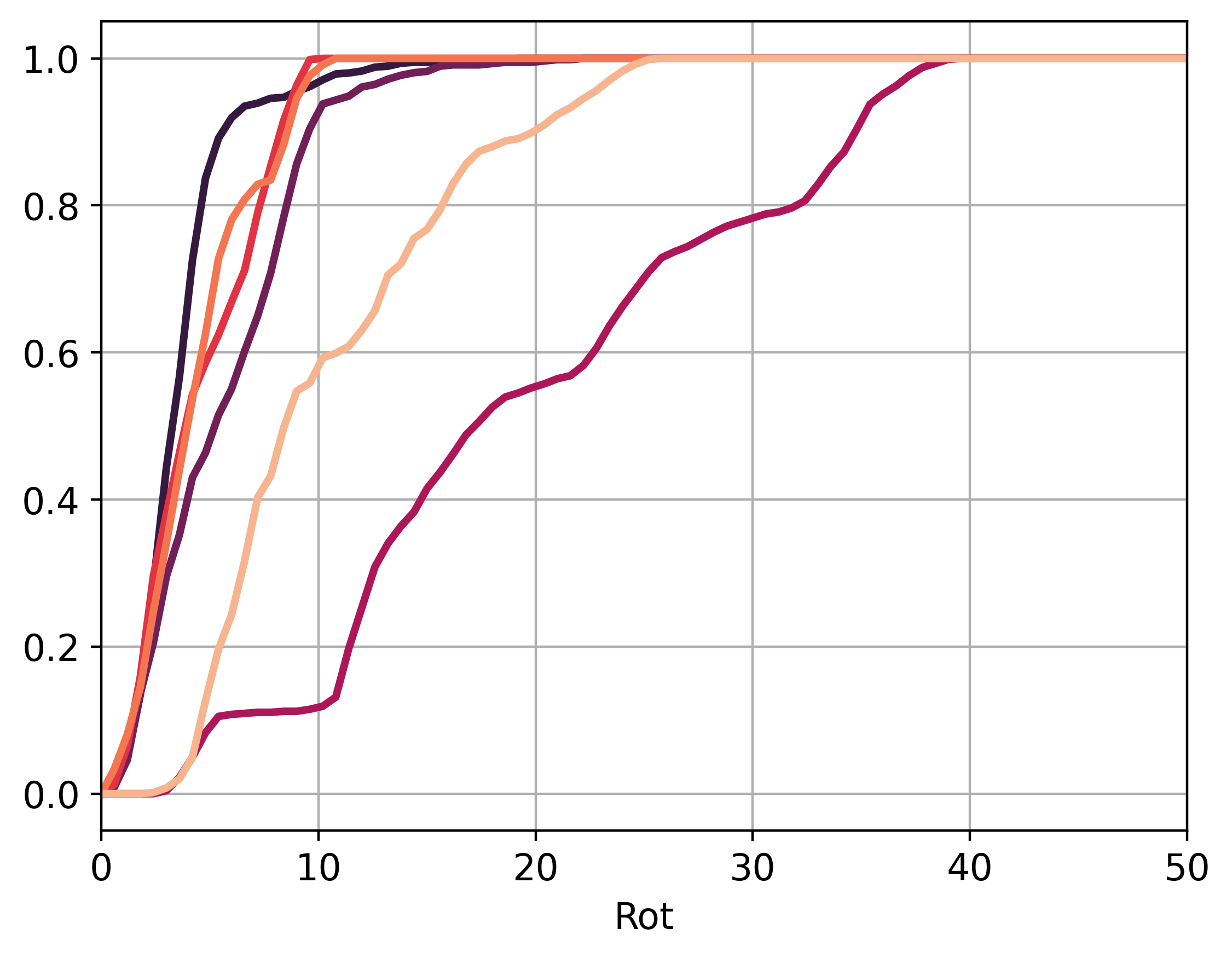}
    \end{subfigure}
    \begin{subfigure}[b]{0.15\textwidth}
    \includegraphics[scale=0.18]{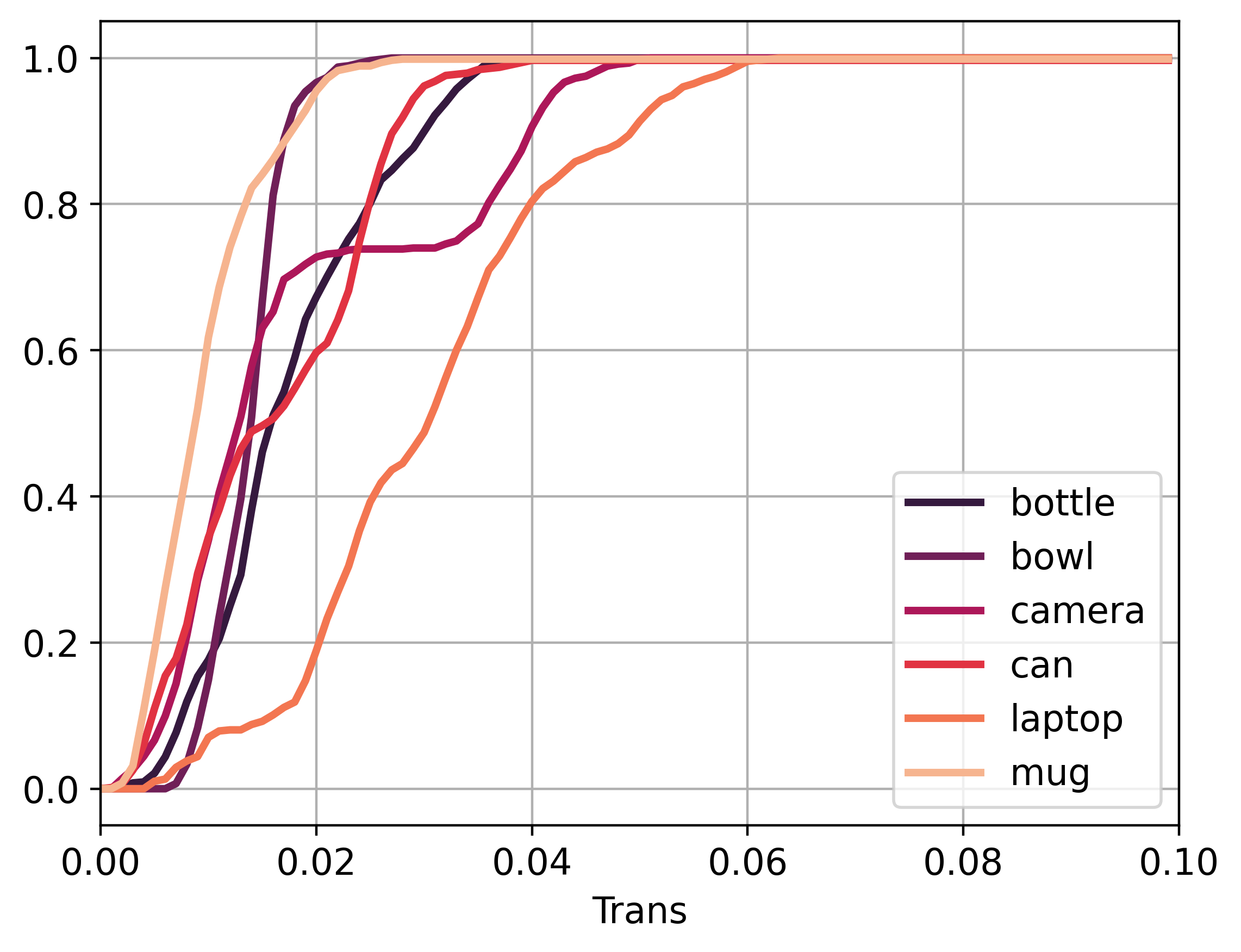}
    \end{subfigure}
    \caption{The AP curve of our approach, with the vertical axis representing AP and the horizontal axis representing IoU, rotation error, and translation error of each category.}
    \label{fig:ap_curve}
\end{figure}

In our experimental results, we present the mean Average Precision (mAP) based on the 3D Intersection over Union (3D IoU) as well as translation and rotation errors, as shown in Table~\ref{table:quantitative_real275}. Figure~\ref{fig:qualitativeResultsREAL} indicates qualitative results. Figure~\ref{fig:ap_curve} shows its AP curve.

On the REAL dataset which contains high-quality depth maps, our method achieves comparable performance to RGB-D methods across various metrics. Specifically, our method shows significant improvements in IoU25 and IoU50 metrics, highlighting the robustness of our approach. The latter stems from three main factors. Firstly, better camera pose and depth estimation can be obtained by utilizing multi-view images. Secondly, by integrating object pose information from multiple frames, we reduce errors and improve the accuracy of pose estimation. Finally, the utilization of multi-view constraints helps mitigate the negative impact of occlusions and false detections that can occur in single view images. This combination of factors contributes to the overall robustness of our method. 
Furthermore, it should be noted that the reason for the lower IoU75 and rotation accuracy compared to RGB-D based methods is due to the significant intra-class variation in the camera category, where the quality of NOCS maps is often insufficient.

\begin{table}[bp]
    \centering
    \resizebox{0.46\textwidth}{!}{%
        \begin{tabular}{cccccc}
            \toprule
            \textbf{Method} & {\textbf{mIoU}(\%)} & {\textbf{mE\textsubscript{rot}}(\degree)} & {\textbf{mE\textsubscript{trans}}(cm)} \\ \cmidrule(lr){1-1}\cmidrule(lr){2-4}
            VI-Net\cite{lin2023vinet} & {42.3 / 59.0 / 79.2} & {48.0 / 36.0 / 12.4} & {36.6 / 15.7 / 8.5} \\
            IST-Net\cite{liu2023istnet} & {48.3 / 55.7 / \textbf{80.8}} & {54.2 / 49.1 / \textbf{10.7}} & {13.5 / 6.9 / 8.6} \\
            \cmidrule(lr){1-4}
            \textbf{Ours} & {\textbf{72.5} / \textbf{75.2} / 78.5} & {\textbf{15.2} / \textbf{14.4} / 12.9} & {\textbf{8.3} / \textbf{3.6} / \textbf{6.6}} \\
         
            \toprule
            
        \end{tabular}
    }
    \caption{mIoU, mE\textsubscript{rot}, mE\textsubscript{trans} respectively denote the mean IoU, rotation and translation errors. The three numbers within the same cell each represent the results in sequences captured by the ZED, Azure Kinect, and the Structure Sensor.} 
    \label{table:mereal_quantativie}
\end{table}

\subsection{Results on MEREAL}
\label{sec:results_MEREAL}
On the MEREAL dataset, we adopt mean IoU and mean translation and rotation error as our comparison criteria. Due to the arbitrary definition of ground truth isotropic scale, we do not include the scale factor when calculating 3D IoUs.
The benchmark results are given by Table~\ref{table:mereal_quantativie}. Moreover, we tested the performance of the proposed method that uses RGB images and IMU as output on the Azure Kinect sequence. With the use of visual-inertial input, the three metrics of the proposed method are respectively 75.0\%, 15.9 degrees, and 6.3 cm, on the same level as that of the RGB-D modality. As the quality of the depth map decreases, the accuracy of single view RGB-D based methods shows a significant decline. However, the proposed method is far less sensitive to this. This occurs because the training depth data of previous methods has been overly idealistic, leading to a significant domain gap with the real world, while our lightweight object pose predictor does not rely on depth and can achieve reliable depth estimation through multi-view images. These factors ultimately enable the proposed method to achieve robust object pose and size estimations.

\begin{figure}[tbp]
\centering
    \begin{subfigure}[b]{0.155\textwidth}
        \includegraphics[width=\linewidth]{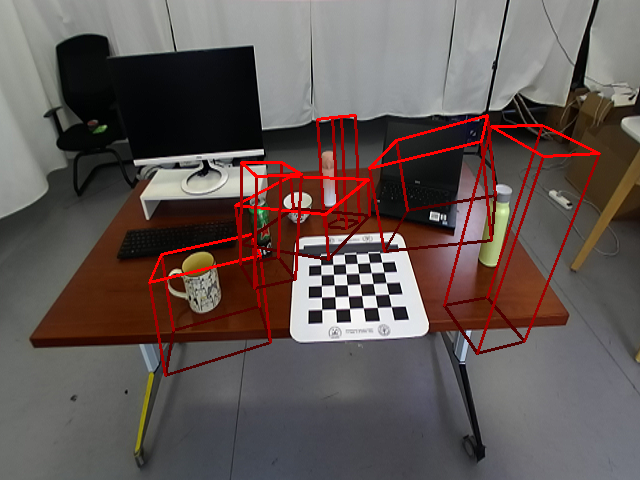}
    \end{subfigure}
    \begin{subfigure}[b]{0.155\textwidth}
        \includegraphics[width=\linewidth]{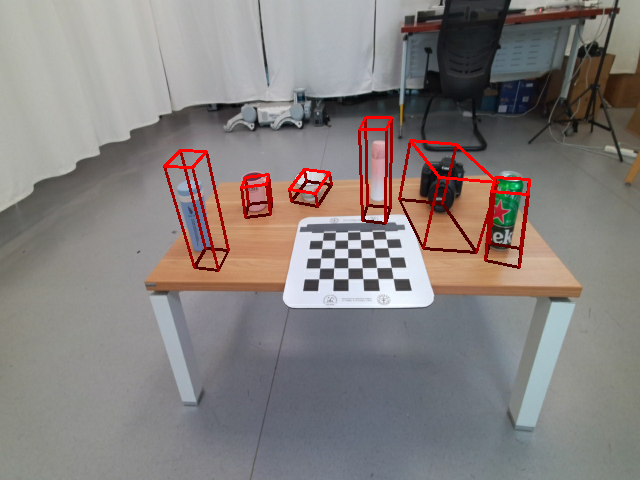}
    \end{subfigure}
    \begin{subfigure}[b]{0.155\textwidth}
        \includegraphics[width=\linewidth]{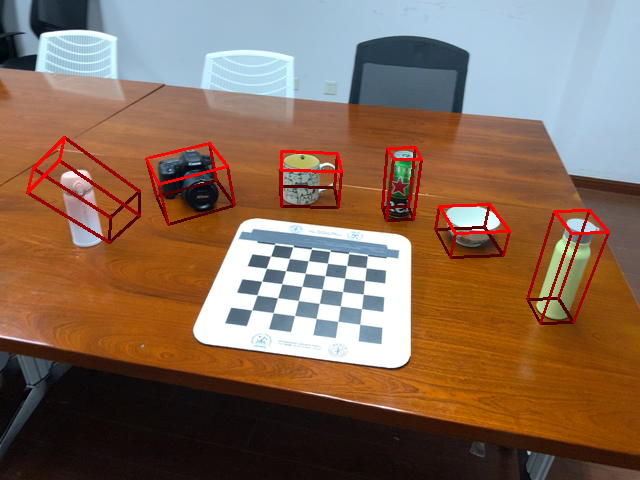}
    \end{subfigure}
    \begin{subfigure}[b]{0.155\textwidth}
        \includegraphics[width=\linewidth]{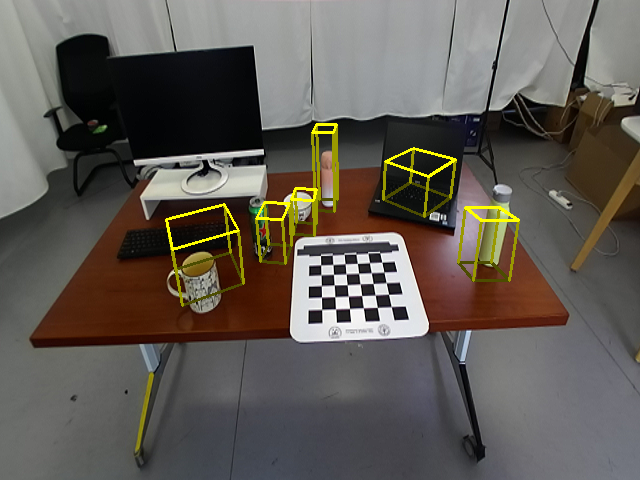}
    \end{subfigure}
    \begin{subfigure}[b]{0.155\textwidth}
        \includegraphics[width=\linewidth]{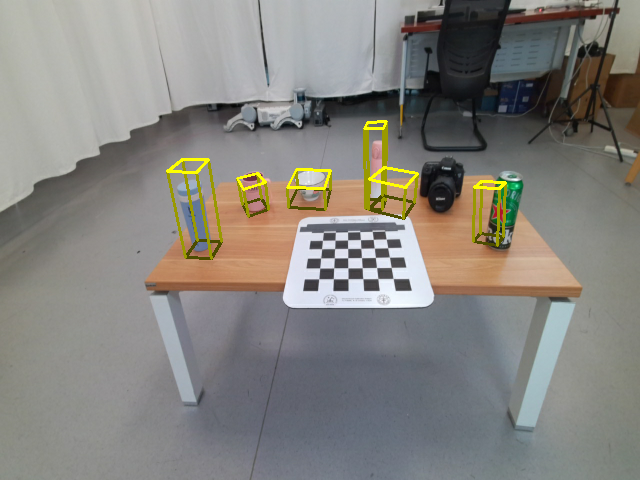}
    \end{subfigure}
    \begin{subfigure}[b]{0.155\textwidth}
        \includegraphics[width=\linewidth]{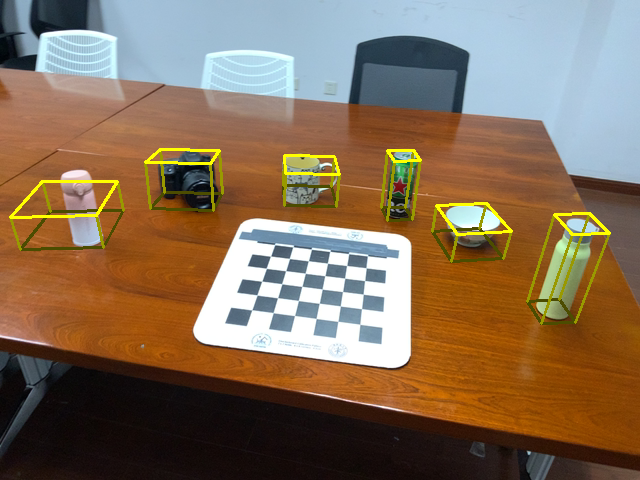}
    \end{subfigure}
    \begin{subfigure}[b]{0.155\textwidth}
        \includegraphics[width=\linewidth]{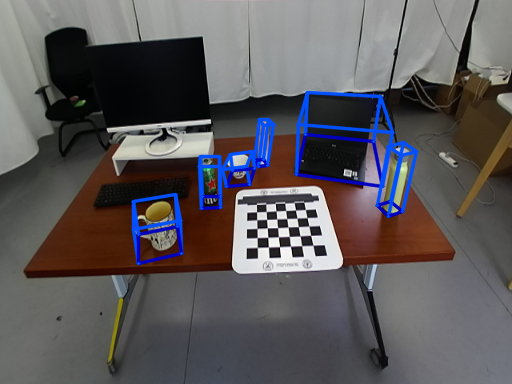}
    \end{subfigure}
    \begin{subfigure}[b]{0.155\textwidth}
        \includegraphics[width=\linewidth]{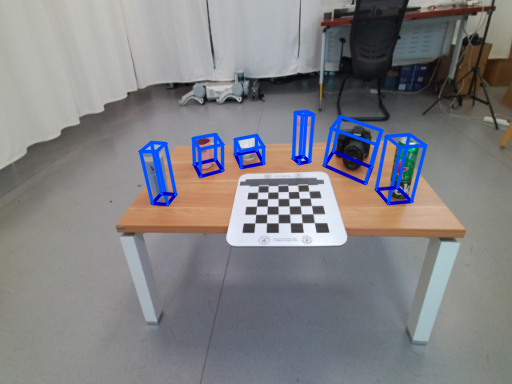}
    \end{subfigure}
    \begin{subfigure}[b]{0.155\textwidth}
        \includegraphics[width=\linewidth]{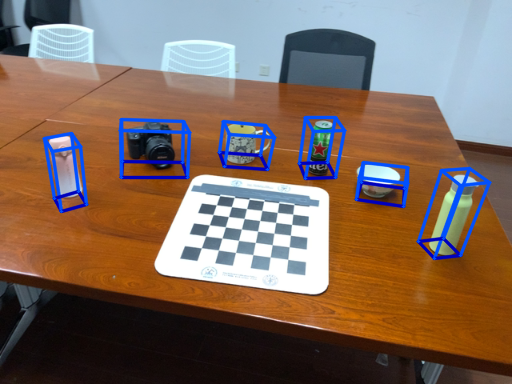}
    \end{subfigure}

    \begin{subfigure}[b]{0.155\textwidth}
        
        \includegraphics[width=\linewidth]{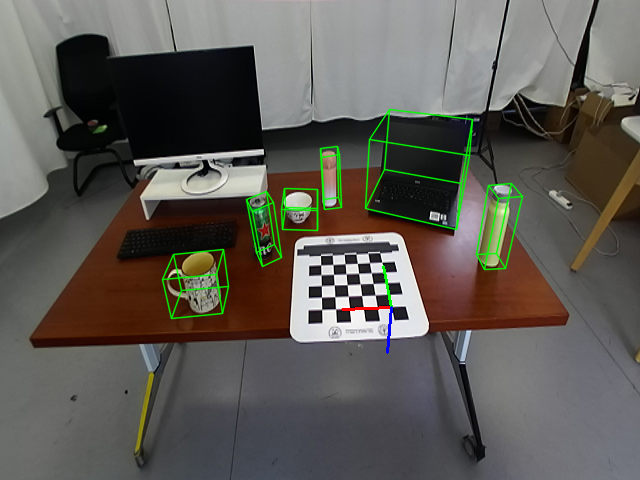}
    \end{subfigure}
    \begin{subfigure}[b]{0.155\textwidth}
        \includegraphics[width=\linewidth]{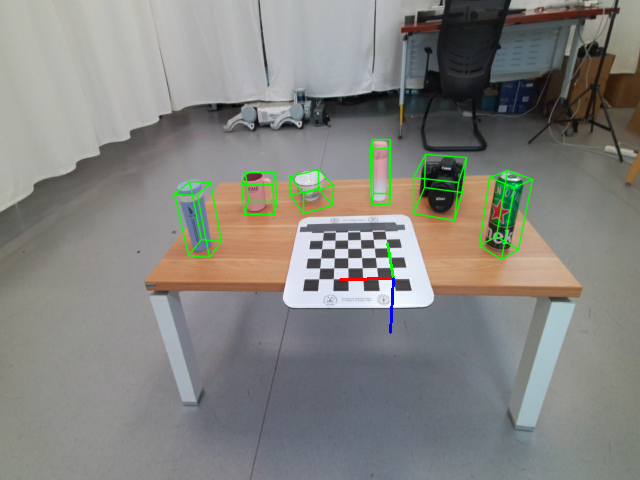}
        
    \end{subfigure}
    \begin{subfigure}[b]{0.155\textwidth}
        \includegraphics[width=\linewidth]{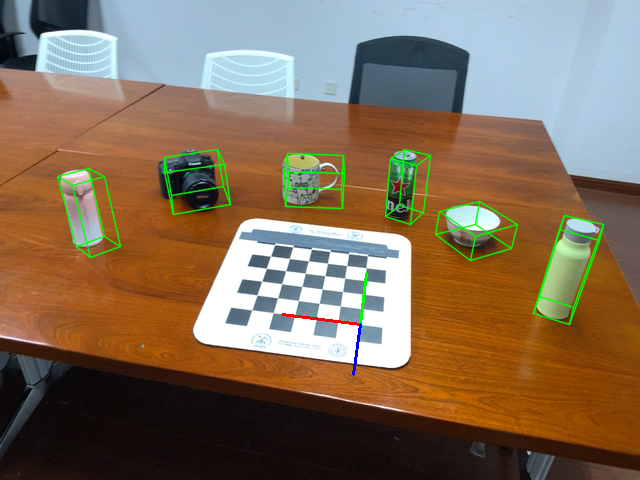}
        
    \end{subfigure}
    \caption{Qualitative results on MEREAL. The four rows, respectively, are: VI-Net~\cite{lin2023vinet}, IST-Net~\cite{liu2023istnet}, the proposed method, and ground truth. The three columns represent the ZED, Azure Kinect, and Structure Sensor sequence, respectively.}
    \label{fig:qualitativeResultsMEREAL}
\end{figure}   

%% file: sec/5_conclusion.tex
\section{Conclusions}
\label{sec:conclusions}

The present work introduces a novel category-level object pose and size estimation framework that distinguishes itself from previous approaches by leveraging a continuous stream of images in conjunction with varying types of scale input. 
It effectively reduces reliance on accurate but expensive depth sensors, and offers multiple flexible ways for deployment in real-world scenarios.
We believe that our method will be of strong interest in applications such as robotic manipulation and virtual reality. 
Our continued efforts consist of utilizing the geometric information from metric predicted depth to enhance the NOCS prediction, and then intensify the pose and size estimation of objects with large intra-class variations.